\documentclass[10pt, conference, letter]{IEEEtran} 

\usepackage{color}
\usepackage{epic, eepic}
\usepackage{graphicx}
\usepackage[vmargin={0.8in,0.8in}, hmargin={0.8in,0.8in}]{geometry}
\usepackage{cite}

\usepackage{amsmath,amssymb}

\usepackage[tight,footnotesize]{subfigure}

\usepackage{url}
\usepackage{balance}

\makeatletter

\newcommand{\Rmnum}[1]{\expandafter\@slowromancap\romannumeral #1@}
\makeatother

\begin{document}
\title{General Scaled Support Vector Machines}

\author{\IEEEauthorblockN{Xin Liu and Ying Ding}
\IEEEauthorblockA{ECHO Labs\\
Nanjing University of Posts and Telecommunications\\
Nanjing, China 210003\\
Email: liuxin.nupt@gmail.com}
\and
\IEEEauthorblockN{Forrest Sheng Bao}
\IEEEauthorblockA{Department of Computer Science/Electrical Engineering\\
Texas Tech University\\
Lubbock, Texas 79401\\
Email: forrest.bao@gmail.com}
}
\maketitle

\begin{abstract}
Support Vector Machines (SVMs) are popular tools for data mining tasks such as classification,
regression, and density estimation. However, original SVM (C-SVM) only considers local information
of data points on or over the margin. Therefore, C-SVM loses robustness. To solve this problem, one
approach is to translate (i.e., to move without rotation or change of shape) the hyperplane according to the distribution of the entire data. But existing
work can only be applied for 1-D case. In this paper, we propose a simple and efficient method
called General Scaled SVM~(GS-SVM) to extend the existing approach to multi-dimensional case.
Our method translates the hyperplane according to the distribution of data projected on the normal
vector of the hyperplane. Compared with C-SVM, GS-SVM has better performance on several data sets.
\end{abstract}


\section{Introduction}
In past several decades, large margin machines have been widely studied and used. Support vector
machines~(SVMs) (also known as C-SVM)~\cite{svmintro}, the most important and effcient one proposed
by Vapnik \textit{et al.}~\cite{statistical}, have been proven of good performance in text mining,
bioinformatics, computer vision, and so forth~\cite{ABG,Noble,Veropoulos}. Unlike many other
classifiers minimizing the \emph{empirical risk}, C-SVM is based on statistical learning 
theory~\cite{statistical}, which emphasizes on minimizing the \emph{structural risk}. C-SVM
constructs a maximal margin between two classes. A hyperplane falls in the middle of this
margin.

While the margin is solely determined by a few data points known as support vectors, remaining data
points have no influence on building the classifier. Obviously, C-SVM loses some robustness because
it cannot use the global information in the entire data set.

Inspired by this observation, we believe it is necessary to embed the global information into
C-SVM. For a binary classification task, the distribution of two classes are usually not the same.
It is reasonable to translate (i.e., to move without rotation or change of shape) the hyperplane closer to the class of the smaller variance.
In~\cite{Feng99}, Feng proposed Scaled SVM (S-SVM) and gave a theoretical distance of the by extreme
theory in 1-D case. 

In this paper, we propose a simple method called General Scaled Support Vector Machine (GS-SVM)
to generalize Feng's method to multi-dimensional case. Our method has three steps. First, it uses
C-SVM algorithms to obtain the hyperplane. Then it projects all data points onto the normal
vector of the hyperplane and estimates the distribution of each class on this direction. Finally, it
translates the hyperplane according to Feng's conclusion. With kernel tricks, we can easily extend our
method to feature space. In this framework, GS-SVM considers both local information of
the data (SVs) and the global information.

The rest of the paper is organized as follows. In the next section, we give a brief background of
C-SVM and Feng's conclusion (S-SVM) that our method bases on. We extend 1-D S-SVM, to
multi-dimensional case, GS-SVM, in Sect.~\ref{sec:our}. Following that, we evaluate
GS-SVM on toy data sets and several benchmarks. This paper is concluded in
Sect.~\ref{sec:conclusion}.

\section{Background}
\subsection{Support Vector Machines}
Support Vector Machines are the implementations of Statistical Learning Theory~\cite{statistical}
which emphasizes on minimizing structural risk. For a binary classfication problem, the two classes
are labeled as $+1$ and $-1$ respectively. The C-SVM problem can be written as:
\begin{equation}
\begin{split}
\min & \| \mathbf{w} \cdot \mathbf{w} \| + C\sum_{i=1}^{l}\xi_{i}^{2}\\
s.t. & \left\{\begin{array}{lc}
y_{i}(\mathbf{w}\cdot \mathbf{x}_{i}+b)\geq 1 - \xi_{i},\quad \\
\xi_{i} \geq 0, \quad 
\end{array}\right.  i=1,2\ldots,l
\end{split} 
\label{csvm}
\end{equation}
where $\xi_{i} \in \mathbb{R}$, $\mathbf{x}_i \in \mathbb{R}^n$ and $y_i\in\{+1,-1\}$ are the
slack variable, feature vector and the label of the $i$-th data point respectively, $n\in \{1, 2,
\ldots \}$ is the dimension of feature vectors, $C$ is the penalty coefficient, and $l$ is the
number of data points. To be solved, $\mathbf{w} \in \mathbb{R}^n$ (weighing vector) and $b$ (bias)
determine the direction and offset of the hyperplane, respectively. $\frac{\| \mathbf{w} \cdot
\mathbf{w} \|}{2}$ is known as the margin width. Laying in the middle of the margin, the hyperplane
bears the equation $\mathbf{w} \cdot \mathbf{x}_i + b = \mathbf{0}, \forall i \in 1..l$.

By the method of Lagrange multipliers, Eq.~(\ref{csvm}) is equivalent to:
\begin{equation}
\begin{split}
\frac{\partial L}{\partial \mathbf{w}} & = \mathbf{w} -  \sum_{i=1}^{l} y_{i}\mathbf{x}_{i}\alpha_i
= \mathbf{0} \\
\frac{\partial L}{\partial \pmb{\xi}} & = C\pmb{\xi} - \pmb{\alpha} = \mathbf{0}\\
\frac{\partial L}{\partial b} & = \sum_{i=1}^{l} \alpha_{i}y_{i} = 0 \\
\end{split}
\label{condition:c-svm}
\end{equation}
where $L$ is the Lagrange function of Eq.~(\ref{csvm}), $\pmb{\xi} = [\xi_1, \xi_2, \ldots,
\xi_l]$, $\pmb{\alpha} = [\alpha_1, \alpha_2, \ldots, \alpha_l]$ and $\alpha_i\ge 0$. Feature vector
$x_i$ such that $\alpha_i \not = 0$ is called a support vector.

Eq.~(\ref{csvm}) can be transformed into its dual form, which also allows the use of
kernel tricks:
\begin{equation}
\begin{split}
 \max~ & -\frac{1}{2}\pmb{\alpha}^{T} \mathbf{Q}
\pmb{\alpha}\quad+\quad\frac{1}{2}\boldsymbol{1}^{T}\pmb{\alpha}\\
 s.t.~ & \sum^{l}_{i=1} y_{i}\alpha_{i} = 0\\
& \alpha_{i} \geq 0, i=1, \ldots, l\\
\end{split} 
\label{dual} 
\end{equation}
where $\mathbf{Q}$ is a matrix whose element at $i$-th row and $j$-th column is $Q_{ij} =
y_{i}y_{j}\phi(\mathbf{x}_{i})\cdot\phi(\mathbf{x}_{j})$, and $\phi$ is a function, e.g., linear
or radial basis, of the feature vector. 

\begin{figure}[h]
\setlength{\unitlength}{5cm}
\begin{picture}(2, 0.3)
\thicklines
\put(0,0){\line(1,0){1.6}}
\put(0.4, 0){\circle{0.05}}
\put(0.48, 0){\circle{0.05}}
\put(0.55, 0){\circle{0.05}}
\put(1.05, 0){\circle*{0.05}}
\put(1.18, 0){\circle*{0.05}}
\put(1.36, 0){\circle*{0.05}}
\thinlines
\put(0.8, -0.3){\line(0,1){0.6}}
\dashline{0.028}(0.7, -0.3)(0.7, 0.3)
\put(0.85, -0.25){$s_{1}$}
\put(0.62, -0.25){$s_{2}$}
\put(0.825, 0.03){$\textbf{0}$}
\put(0.25, -0.08){$-b$}
\put(1.55, -0.08){$a$}
\put(0.625, 0){\oval(0.15, 0.1)[t]}
\put(0.875,0){\oval(0.35, 0.1)[b]}
\put(0.625, 0.1){$c_{1}$}
\put(0.875, -0.1){$c_{2}$}
\put(0.48, 0.2){$A_{2}$}
\put(1.18, 0.2){$A_{1}$}
\end{picture}
\vspace{1cm}
\caption{An illustration of Scaled-SVM in 1-D\label{1d-scale}. $s_1$ and $s_2$ are the hyperplanes
obtained by C-SVM and Scaled-SVM respectively. }
\end{figure}
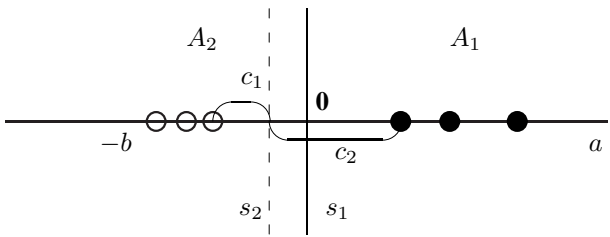

\subsection{Scaled SVM}
Due to the sparseness of $\pmb{\alpha}$, the hyperplane is only related to a few points while other
points have no influence. Therefore, C-SVM loses some robustness. Based on this observation,
Feng~\cite{Feng99} proposed Scaled SVM taking the distribution scale (range) of two classes into
consideration. This method can advance C-SVM by at most $10\%$ on average generalization error.

Assume two classes $A_{1}$ and $A_{2}$ in one dimension are distributed in intervals $(0, a)$ and
$(-b, 0)$ respectively, where $a, b > 0$. Let $s_{1}$ be the hyperplane obtained by C-SVM. Denote
$d_{1}$ and $d_{2}$ as the distribution scales of $A_{1}$ and $A_2$, respectively. Let $c_{1}$ and
$c_{2}$ be the distances from $s_{2}$ to the nearest points in $A_{1}$ and $A_{2}$, respectively.
According to Scaled SVM, in parallel with $s_1$, the new hyperplane $s_{2}$ satisfies:
\begin{equation}
 \frac{c_{1}}{c_{2}} = \sqrt{\frac{d_{1}}{d_{2}}}\label{new}~,
\end{equation}
 as illustrated in Fig.~\ref{1d-scale}.

Eq.~(\ref{new})
can be reformulated into:
\begin{equation}
x_{new} = x_{old} + \Delta = x_{old} +
\frac{2(\sqrt{d_{2}}-\sqrt{d_{1}})}{\sqrt{d_{1}}+\sqrt{d_{2}}}~,
\label{delta}
\end{equation}
where $x_{old}$ and $x_{new}$ are the locations of $s_1$ and $s_2$ respectively. The calculation of
$\Delta$ in multi-dimensional case will be determined later in the paper.

\subsection{Related Work}
There have been many works which aim at combining the global information into C-SVM. Huang
\textit{et al.} proposed a new large margin classifier called Maxi-Min Margin Machine~($M^{4}$)
which use the covariance information of two classes~\cite{Huang}. Yeung~\textit{et al.} first used
clustering algorithms to determine the structure of data, then incorporated this structural
information into constraints to calculate the largest margin~\cite{SLMM}. In contrast to
integrating global information into constraints, Xue~\textit{et al.}~\cite{Xue} proposed Structural
Support Vector Machine, which embeds global information into the C-SVM's objective function. This
approach greatly reduces the computational complexity while keeping the sparsity merit of C-SVM.
Xiong and Cherkassky proposed SVM/LDA which combined LDA and SVM together~\cite{Xiong}. The SVM part
reflects the local information of the
data while the LDA part reflects the global information. Takuya and Shigeo improved the
generalization ability of C-SVM by optimizing the bias term based on Bayesian theory~\cite{Bayes}.

\section{Our Proposed Method}
\label{sec:our}
\subsection{Overview}
\begin{figure}[!hbt]
\includegraphics[scale=0.45]{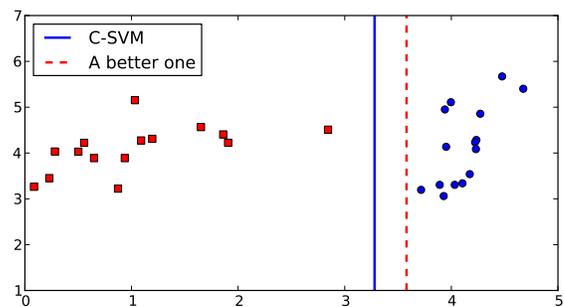}
\caption{An illustration of our idea\label{idea}. The blue solid line is the hyperplane obtained by
C-SVM. The red dash line is a better one for it is closer to the class (blue circles) with smaller
variance on the horizontal direction.}
\end{figure}

To a binary classification task, C-SVM will put the hyperplane in the middle of the margin. However,
since the distributions of two classes are usually different, it makes sense
 to translate the hyperplane away from the class of larger variance and toward the other class. An illustration is shown in
Fig.~\ref{idea}.

Let the translating distance of the hyperplane be $\Delta$. The new SVM is the solution of the 
optimization problem below:

\begin{equation}
\begin{split}
\min\quad & \| \mathbf{w} \cdot \mathbf{w}\| + C\sum_{i=1}^{l}\xi_{i}^{2}\\
s.t.\quad & \left\{\begin{array}{lc}
y_{i}(\mathbf{w}\cdot \mathbf{x}_{i}+b)\geq 1+\Delta-\xi_{i}, \text{~if~} y_i = +1 \\
y_{j}(\mathbf{w}\cdot \mathbf{x}_{j}+b)\geq 1-\Delta-\xi_{j}, \text{~if~} y_j = -1\\
\end{array}\right.
\end{split}
\label{opt}
\end{equation}

Without losing of generalization, $i=1,2,\ldots,l^{+}$, $j=l^{+}+1,\ldots,l$, and $l^+$ is total number of positive class. 

The solution of Eq.~(\ref{opt}) is called a \emph{General Scaled SVM}.

The Lagrange function of Eq.~(\ref{opt}) is:
\begin{eqnarray*}
 L &=& \frac{1}{2}\| \mathbf{w}\cdot \mathbf{w}\| +
\frac{C}{2}\sum_{i=1}^{l}\xi_{i}^{2} \\ 
	      &-& \sum_{i=1}^{l^{+}}\alpha_{i}[y_{i}(\mathbf{w}\cdot
\mathbf{x}_{i}+b)-1-\Delta+\xi_{i}] \\
	     &-& \sum_{j=l^{+}+1}^{l}\alpha_{j}[y_{j}(\mathbf{w}\cdot
\mathbf{x}_{j}+b)-1+\Delta+\xi_{j}] 
\end{eqnarray*}

Eq.~(\ref{opt}) is equivalent to 
\begin{equation}
\begin{split}
\frac{\partial L}{\partial \mathbf{w}} 
& =  \mathbf{w} - \sum_{i=1}^{l} y_{i}\alpha_{i}\mathbf{x}_{i}
+ \frac{\partial \Delta}{\partial \mathbf{w}} \left [ \sum_{i=1}^{l^{+}}\alpha_{i}-\sum_{j=l^{+}+1}^{l}\alpha_{j}\right ]\\
& = \mathbf{w} - \sum_{i=1}^{l} y_{i}\alpha_{i}\mathbf{x}_{i}
+ \frac{\partial \Delta}{\partial \mathbf{w}} \sum_{i=1}^{l}\alpha_{i}y_i\\
& = \mathbf{w} -  \sum_{i=1}^{l} y_{i}\alpha_{i}\mathbf{x}_{i} = \mathbf{0} \\
\frac{\partial L}{\partial \pmb{\xi}} & = C\pmb{\xi} - \pmb{\alpha} = \mathbf{0}\\
\frac{\partial L}{\partial b} & = \sum_{i=1}^{l} \alpha_{i}y_{i} = 0 \\
\end{split}
\label{condition:gs-svm}
\end{equation}

Since Eq.~(\ref{condition:gs-svm}) is identical to Eq.~(\ref{condition:c-svm}) (of C-SVM) and $L$ is
independent from $\Delta$, this problem can be solved in three steps:

\begin{enumerate}
\item Use the C-SVM algorithm to obtain the original hyperplane.
\item Project all the points onto the normal vector of the hyperplane and estimate the distribution
of each class in the projection.
\item Calculate $\Delta$ and translate the original hyperplane to obtain the new one.
\end{enumerate}
\vspace{0.6cm}
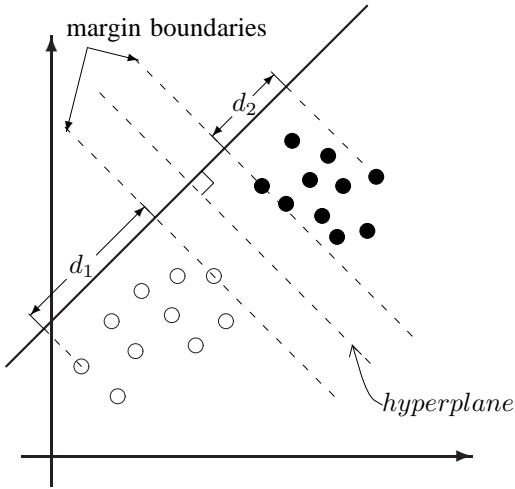
\begin{figure}[!hbt]
\centering
\setlength{\unitlength}{4cm}
\begin{picture}(1.2, 0)
\thicklines
\put(0, -1.5){\vector(0,1){1.5}}
\put(-0.1, -1.4){\vector(1,0){1.5}}
\put(-0.15, -1.1){\line(1,1){1.2}}
\thinlines
\put(0.1, -1.1){\circle{0.05}}
\put(0.2, -0.95){\circle{0.05}}
\put(0.28, -1.05){\circle{0.05}}
\put(0.4, -0.93){\circle{0.05}}
\put(0.42, -0.8){\circle{0.05}}
\put(0.48, -1.03){\circle{0.05}}
\put(0.3, -0.85){\circle{0.05}}
\put(0.58, -0.95){\circle{0.05}}
\put(0.54, -0.8){\circle{0.05}}
\put(0.22, -1.2){\circle{0.05}}
\put(0.7, -0.5){\circle*{0.05}}
\put(0.86, -0.48){\circle*{0.05}}
\put(0.95, -0.67){\circle*{0.05}}
\put(0.9, -0.6){\circle*{0.05}}
\put(1.08, -0.47){\circle*{0.05}}
\put(0.97, -0.5){\circle*{0.05}}
\put(0.78, -0.56){\circle*{0.05}}
\put(0.92, -0.4){\circle*{0.05}}
\put(0.8, -0.35){\circle*{0.05}}
\put(1.05, -0.65){\circle*{0.05}}
\dashline{0.028}(0.94, -1.2)(0.045, -0.305) 
\dashline{0.028}(1.2, -1)(0.275, -0.075)
\dashline{0.028}(1.06, -1.09)(0.16, -0.19)
\dashline{0.028}(0.1, -1.1)(-0.025, -0.975)  
\dashline{0.028}(0.78, -0.17)(1.08, -0.47)
\put(0.05, 0){margin boundaries}
\put(0.12, -0.04){\vector(-1, -4){0.07}}
\put(0.12, -0.04){\vector(4, -1){0.17}}
\put(1.1, -1.25){$hyperplane$}
\spline(1.1, -1.23)(1.05, -1.2)(1.0, -1.03)
\put(1.0, -1.03){\line(-1, -4){0.01}}
\put(1.0, -1.03){\line(1, -1){0.03}}
\put(0.5, -0.53){\line(1,1){0.04}}
\put(0.5, -0.45){\line(1,-1){0.04}}
\put(0.06, -0.79){$d_1$}
\put(0.6, -0.25){$d_2$}
\put(-0.025, -0.975){\line(-1, 1){0.05}}
\put(0.345, -0.605){\line(-1, 1){0.05}}
\put(0.575, -0.375){\line(-1, 1){0.05}}
\put(0.78, -0.17){\line(-1, 1){0.05}}
\put(0.05, -0.81){\vector(-1,-1){0.12}}
\put(0.15, -0.73){\vector(1,1){0.16}}
\put(0.6, -0.27){\vector(-1,-1){0.065}}
\put(0.67, -0.2){\vector(1, 1){0.07}}
\end{picture}
\vspace{6cm}
\caption{An illustration of calculating $\Delta$.\label{cal_delta}}
\end{figure}

\subsection{Calculating	 $\Delta$}
After obtaining $\mathbf{w}$ from C-SVM training, we project data points onto the normal vector of 
the hyperplane.
Then we utilize the projected scale of each class  to adjust the hyperplane. This is
illustrated in Fig.~\ref{cal_delta}. Feng's conclusion
for 1-D SVMs can be extended to multi-dimensional case.

In the input space, the projected coordinate of any data point $\mathbf{x}_{i}$ on the normal vector of the hyperplane
is:
\begin{equation}
e_{i} = \frac{\mathbf{w}\cdot \mathbf{x}_{i}}{\|\mathbf{w}\|}
\end{equation}

Projected scales can be calculated as:
\begin{eqnarray}
d_{1} &=& \max \mathbf{E}_{+} - \min \mathbf{E}_{+} \nonumber \\
d_{2} &=& \max \mathbf{E}_{-} - \min \mathbf{E}_{-} \nonumber 
\end{eqnarray}
where $\mathbf{E}_{+} = \{e_i | y_i=+1\}$ and $\mathbf{E}_{-} = \{ e_i| y_i=-1\} $.

With $d_1$ and $d_2$, $\Delta$ can be calculated as in Eq.~(\ref{delta}):
$$\Delta = \frac{2(\sqrt{d_{2}}-\sqrt{d_{1}})}{\sqrt{d_{1}}+\sqrt{d_{2}}}$$
 Note that $\Delta$ ranges from $-1$ to $1$.

In feature space, where all data points $\mathbf{x}$ are mapped into $\phi(\mathbf{x})$, since
$\mathbf{w} = \sum_{i=1}^{l}\alpha_{i}\phi(\mathbf{x}_{i})$, all we need to do is replacing $\langle
\mathbf{x}_{i}\cdot \mathbf{x}_{j}\rangle$ with $K(\mathbf{x}_{i}, \mathbf{x}_{j})$.

\begin{figure*}[!hbt]
\centering
\subfigure[training session]{\label{train}
\hskip -1.7em
\includegraphics[scale=0.43]{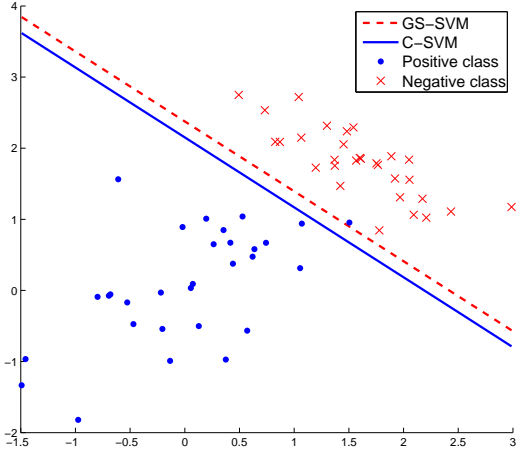}}
\subfigure[testing session]{\label{test}
\includegraphics[scale=0.43]{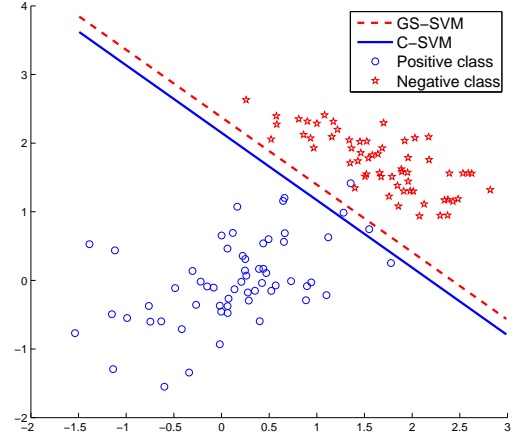}}
\caption{An illustration of toy data\label{toy}.}
\end{figure*}

\section{Experiments}
\label{sec:exp}
In this section, we first demonstrate the advantage of GS-SVM on synthetic 2-D toy data sets. Then
we compare GS-SVM with C-SVM on several real world 
benchmarks. The training and testing of SVMs are accomplished by LIBSVM~\cite{CC01a}.
\subsection{2-D Toy Data}
As illustrated in Fig.~\ref{train}, the data set is generated under two Gaussian distributions: the positive class is randomly sampled from the Gaussian distribution with the mean as
$[0.2, 0.1]^{T}$ and the covariance as $[0.5,0.2;0.2,0.4]$, while the negative class is randomly sampled from another distribution with the mean and the covariance as 
$[1.7,1.7]^{T}$ and $[0.4,-0.2;-0.2,0.4]$. Training and test sets consist of 30 and 60 data points respectively for each class. Fig.~\ref{test} illustrates the hyperplanes derived by C-SVM and GS-SVM.
From Fig.~\ref{test}, we find that GS-SVM achieves a better hyperplane by taking both the local and global information of the data into consideration when determining the 
position of the hyperplane. As expected, the GS-SVM translates the hyperplane toward the class (negative class) of smaller projected scale on the normal of the hyperplane. GS-SVM classifies two more points correctly. The classification accuracies 
of C-SVM and GS-SVM are $96.67\%$ and $97.5\%$ respectively. The improvement on accuracy demonstrates the advantage of our proposed method.

\subsection{Benchmarks}
We also evaluate GS-SVM on 8 standard data sets from UCI machine learning repository~\cite{Frank}.
GS-SVM are compared with 
C-SVM on both the linear and Gaussian kernels. The parameter $C$ for both methods is tuned via
10-fold cross validation. So is the width
parameter of Gaussian kernel. The performance of these two methods in 10-fold cross validation is summarized in Table~\ref{result}.

\begin{table}
\centering
\begin{tabular}{|l|c|c||c|c|}
\hline
data sets& \multicolumn{2}{c||}{linear kernel}&\multicolumn{2}{c|}{Gaussian kernel}\\ \cline{2-5}
& C-SVM & GS-SVM & C-SVM & GS-SVM\\ \hline
sonar &73.72  &\textbf{75.10} &88.47 &\textbf{89.87} \\ \hline 
liver &68.28 &\textbf{69.81} &73.91 &\textbf{74.5} \\ \hline
heart & 83.33& \textbf{83.71}& 83.33&\textbf{84.44} \\ \hline
spect & 76.47&\textbf{77.01}&89.03 &\textbf{89.84} \\ \hline
breast & 96.81& 96.81&97.22 & \textbf{97.36}\\ \hline
statlog & 84.95& 84.95& 86.37&\textbf{86.95} \\ \hline
diabet &76.95 &\textbf{77.34} &77.86 &\textbf{78.26} \\ \hline
hepatitis &78.28 &\textbf{80.39}& 83.28&\textbf{84.54} \\ \hline
\end{tabular}
\caption{Comparisons of classification accuracies among C-SVM and GS-SVM~\label{result}}
\end{table}

GS-SVM achieves a better performance on most data sets  in both linear  and Gaussian kernel. On
remaining  data sets, GS-SVM performs as well as C-SVM. The results on these benchmarks show that
it is worth considering the global information of the data. 

\begin{figure}
\centering
\includegraphics[scale=0.4]{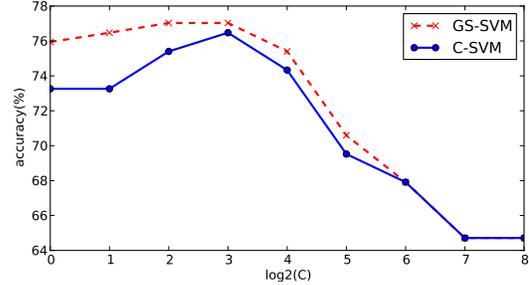}
\caption{Accuracy versus C\label{c_delta} on ``spect'' data set.}
\end{figure}

We notice an interesting 
role that $\Delta$ performs: GS-SVM can reach a higher accuracy with a smaller penalty value $C$
than C-SVM. It is not hard to understand from Eq.~(\ref{opt}).
We select ``spect'' data set as an example and show the relationship between $C$ and the accuracy in
Fig.~\ref{c_delta}. Since the hyperplane translates toward the class of smaller projected scales on the
normal of the hyperplane, 
 it is more possible from the sum of slack variables $\sum_{i=1}^l \xi _i$ to decrease.
$C$ is also used to minimize the classification error.
The greater  $C$ is, the smaller the classification error will be. As the hyperplane translates, 
it is more possible for classification error to drop. Hence, GS-SVM will achieve a better performance with a lower $C$.
Note that although $C$ and $\Delta$ have the same effect of adjusting the position of the
hyperplane, they do not work in the same way. $C$ adjusts the margin (the hyperplane lays in the
middle of the margin) to minimize the training error, while $\Delta$ scales the position of the
hyperplane to minimize structural risk. 

\section{Conclusion}
\label{sec:conclusion}
In this paper, we propose a simple but efficient method to improve the generalization ability of C-SVM, called as GS-SVM. C-SVM only uses 
support vectors and ignores the information of other data points. Previous works have been done to consider global information 
in deciding the hyperplane. For binary classification problem, one approach is to translate the
hyperplane toward the class with smaller projected scale on the direction that is perpendicular to
the hyperplane. However, existing work of this approach is only for 1-D case. In this paper, this
approach is extended from 1-D to multi-dimensional cases. Experimental results show that GS-SVM
advances C-SVM on both toy data sets and most of the benchmarks used. Throughout the paper, we
discuss our method in the binary classification problem. However, it can be easily extended to
multi-class classification problem. A future investigation will focus on theoretical analysis on the
generalization ability of GS-SVM.

\balance{}
\bibliographystyle{IEEEtran}
\bibliography{1}
\end{document}